\documentclass{sig-alternate-05-2015}

\begin{document}

% Copyright
\setcopyright{acmcopyright}
%\setcopyright{acmlicensed}
%\setcopyright{rightsretained}
%\setcopyright{usgov}
%\setcopyright{usgovmixed}
%\setcopyright{cagov}
%\setcopyright{cagovmixed}

% DOI
\doi{**.***/***_*}

% ISBN
\isbn{***-****-**-***/**/**}

%Conference
\conferenceinfo{ICMR '16}{June 6--9, 2016, New York, NY, USA}

\acmPrice{\$15.00}

%
% --- Author Metadata here ---
%\conferenceinfo{WOODSTOCK}{'97 El Paso, Texas USA}
%\CopyrightYear{2007} % Allows default copyright year (20XX) to be over-ridden - IF NEED BE.
%\crdata{0-12345-67-8/90/01}  % Allows default copyright data (0-89791-88-6/97/05) to be over-ridden - IF NEED BE.
% --- End of Author Metadata ---

\title{GPU-accelerated Hierarchical Panoramic Image Feature Retrieval for Indoor Localization}
%\subtitle{[Extended Abstract]
%\titlenote{A full version of this paper is available as
%\textit{Author's Guide to Preparing ACM SIG Proceedings Using
%\LaTeX$2_\epsilon$\ and BibTeX} at

\numberofauthors{3} %  in this sample file, there are a *total*
\author{
% 1st. author
\alignauthor
Feng Hu\\
       \affaddr{The Graduate Center,}\\
       \affaddr{ CUNY}\\
       \affaddr{365 5th Ave, New York}\\
       \affaddr{New York, United States}\\
       \email{fhu@gradcenter.cuny.edu}
% 2nd. author
\alignauthor
Zhigang Zhu\\
       \affaddr{The Graduate Center and City College, CUNY}\\
       \affaddr{138th St \& Convent Ave}\\
       \affaddr{New York, United States}\\
       \email{zzhu@ccny.cuny.edu }
% 3rd. author
\alignauthor 
Jianting Zhang\\
       \affaddr{The Graduate Center and City College, CUNY}\\
       \affaddr{138th St \& Convent Ave}\\
       \affaddr{New York, United States}\\
       \email{jzhang@cs.ccny.cuny.edu }
%\and  % use '\and' if you need 'another row' of author names
% 4th. author
%\alignauthor Lawrence P. Leipuner\\
%       \affaddr{Brookhaven Laboratories}\\
%       \affaddr{Brookhaven National Lab}\\
%       \affaddr{P.O. Box 5000}\\
%       \email{lleipuner@researchlabs.org}
% 5th. author
%\alignauthor Sean Fogarty\\
%       \affaddr{NASA Ames Research Center}\\
%       \affaddr{Moffett Field}\\
%       \affaddr{California 94035}\\
%       \email{fogartys@amesres.org}
% 6th. author
%\alignauthor Charles Palmer\\
%       \affaddr{Palmer Research Laboratories}\\
%       \affaddr{8600 Datapoint Drive}\\
%       \affaddr{San Antonio, Texas 78229}\\
%       \email{cpalmer@prl.com}
}
% There's nothing stopping you putting the seventh, eighth, etc.
% author on the opening page (as the 'third row') but we ask,
% for aesthetic reasons that you place these 'additional authors'
% in the \additional authors block, viz.
%\additionalauthors{Additional authors: John Smith (The Th{\o}rv{\"a}ld Group,
%email: {\texttt{jsmith@affiliation.org}}) and Julius P.~Kumquat
%(The Kumquat Consortium, email: {\texttt{jpkumquat@consortium.net}}).}
% Just remember to make sure that the TOTAL number of authors
% is the number that will appear on the first page PLUS the
% number that will appear in the \additionalauthors section.

\maketitle
\begin{abstract}
Indoor localization has many applications, such as commercial Location Based Services (LBS), robotic navigation, and assistive navigation for the blind. This paper formulates the indoor localization problem into a multimedia retrieving problem by  modeling visual landmarks with a panoramic image feature, and calculating a user's location via GPU-accelerated parallel retrieving algorithm.  
%To evaluate the localization accuracy, we map both the testing images' features and the database features onto the ground truth 2D floor plan's coordinates, and measure the localization errors in physical space. 
To solve the  scene similarity problem, we apply a multi-images based retrieval strategy and a 2D aggregation method to estimate the final retrieval location. Experiments on a campus  building real data demonstrate real-time responses (14fps) and robust localization.
\end{abstract}




%
%  Use this command to print the description
%

% We no longer use \terms command


\section{INTRODUCTION}
Localization is an important task for human beings, since people always need to know where they are. 
%all the time where they are everywhere in the world for variant of tasks. 
As a subset problem of the general localization task, indoor localization is becoming a hot research area as more and more Location Based Services (LBS) are emerging. The applications include museum guidance, shopping mall tour guide, and assistive localization for the visually impaired, to name a few.

For the outdoor localization, Global Navigation Satellite Systems (GNSS), such as 
%Global Positioning System (GPS)\footnote{http://www.gps.gov/}, Galileo system\footnote{http://www.gsa.europa.eu/}, Beidou system\footnote{http://en.beidou.gov.cn/}, 
Global Positioning System (GPS), Galileo system, Beidou system, 
are  widely used to provide accurate localization services worldwide, for example, vehicle  navigation, running tracking, and many other smart phone based LBS.  However, due to the significant signal attenuation caused by construction materials and that the microwave scattering by roofs, walls and other objects, using GNSS systems for indoor localization tasks is not practical. 

\begin{figure}[t]
\begin{center}
   \includegraphics[width=\linewidth]{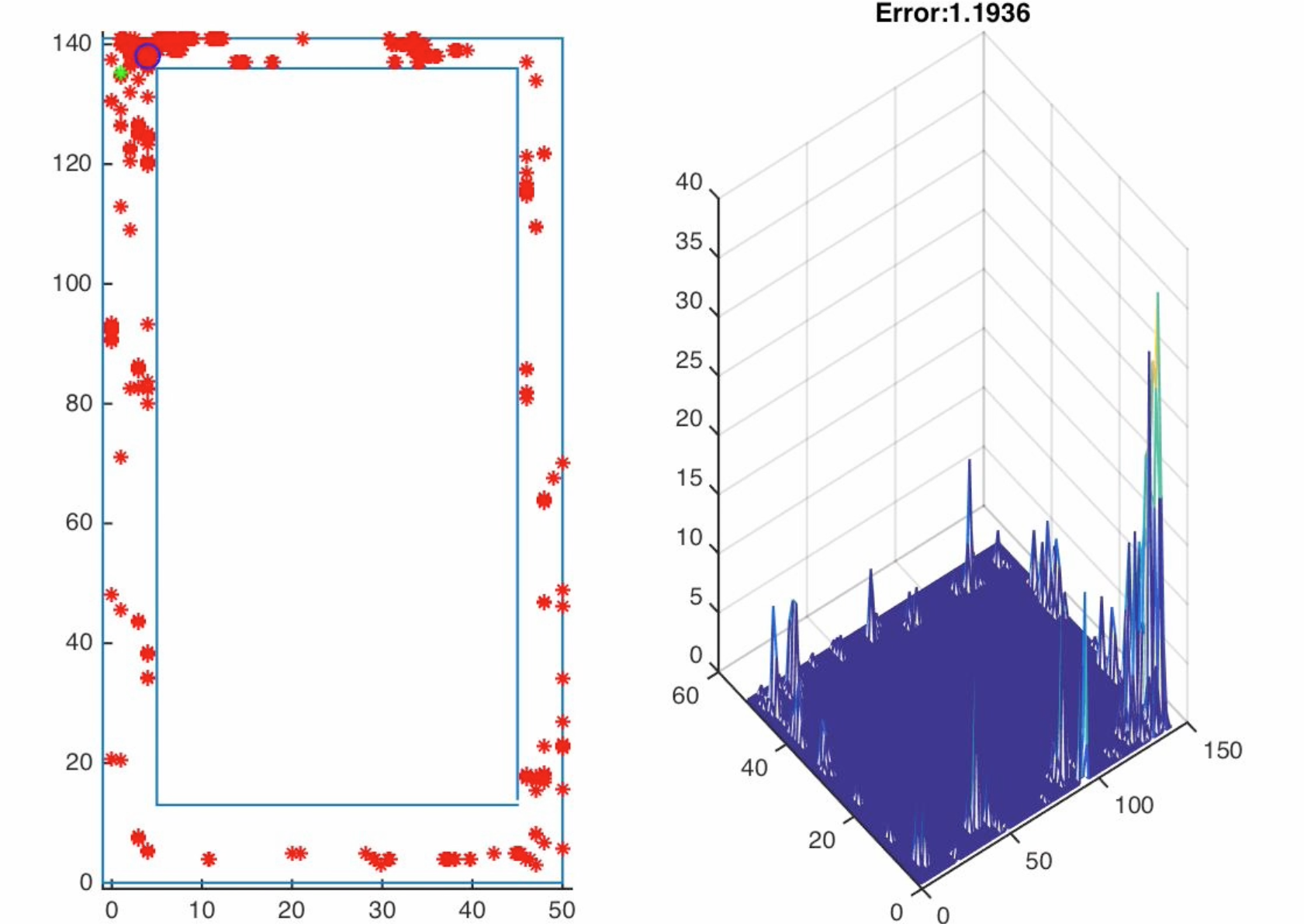}
\end{center}
   \caption{An example of retrieval and aggregation results of a multi-frame  omnidirectional feature of an input on a multi-path databases}
\label{fig:long}
\end{figure}

There are other sensors based localization methods developed for indoor localization in recent years. 
%Radio Frequency Identification Devices(RFID) and Wireless Fidelity(WiFi) 
RFID and WiFi are two frequently used ones. However, RFID-based system, for example,  \cite{cicirelli2012rfid} requires a large number of tags to be installed in the environment before the user can use his/her localization device. WiFi based localization method \cite{paisios2012mobile} can achieve relatively high accuracy, but it depends on the number of access points available, and also cannot be used in an environment where few or no Internet routers are available. 

Visual information based localization is attracting more and more interests in both the computer vision community and the robotics community, since visual information encodes very rich knowledge about the indoor environment. These visual information include all the visible physical objects existing in the environments, for example, doors, windows, or notice boards. Even though the general problem of object recognition is not fully solved, and thus the object recognition based localization method is not mature, there are still many other methods we can utilize to explore these visual information, for example, by extracting primitive geometric elements, such as vertical lines of the environment, to differentiate environmental landmarks and distinguish them for accurate localization.

In this paper, we  propose a GPU-accelerated indoor localization approach, and formulate  the  visual-based indoor  localization problem into a multimedia retrieving problem by dividing it into two stages. First, we model the indoor environment using an omnidirectional image feature proposed by \cite{hu2014mobile}. %For each landmark of the environment, there is a corresponding omnidirectional image feature extracted and stored in the database correspondingly. 
Second, for any given new omnidirectional images captured at an unknown position by the user, the user's location can be real-time calculated by retrieving the most similar image feature from the database via GPU-accelerated  parallel retrieving  and candidate aggregation algorithm. Fig. \ref{fig:long} shows an example of retrieving and aggregation result with an input omnidirectional feature as well as its neighbor features against a multi-path indoor environment databases. The red stars are the retrieved candidates by GPU, the blue circle is the estimated location, and the green star is the ground truth location. Candidate density distribution, which is used to determine the final estimation is also shown on the right picture. Details will be provided in Section \ref{Algorithm}. 

The organization of the rest of the paper is as follows. Section 2 discusses some related work. Section 3 explains the main retrieval and  aggregation algorithms. Section 4 shows the experimental results on data of a real campus building. Section 5 gives a conclusion and points out some possible future directions. 

\section{RELATED WORK}
Direct image retrieval entails retrieving those image(s) in a collection that satisfy a user's need, either using a keyword associated with the image, which forms the category of concept-based image retrieval, or using the contents (e.g. color, shape) of the image, which forms the category of content based image retrieval \cite{kovashka2012whittlesearch}. In our work, we do not directly use images as the retrieving keys, but we extract and use an one dimensional omnidirectional feature from the omnidirectional images.

%% FENG: this is an IR conference so everyone knows this so we can skip them
%Lew\cite{lew2006content}, Rui\cite{rui1999image} and Idris\cite{idris1997review} wrote a few comprehensive review papers for general image retrieval. The earliest image retrieval work, for example, Haralick\cite{haralock1991computer}, focuses on feature based similarity search over images. There are also some early industrial image retrieving systems, for example, QBIC\cite{flickner1995query} and Virage\cite{bach1996virage}. Later on, algorithms are integrated into popular commercial systems such as Informix datablades, IBM DB2 Extenders, or Oracle Cartridges\cite{bliujute1999developing}. 

The applications of content based image retrieval include sketch-based image searching \cite{Jin:2015:NVA:2671188.2749302}, changing environment place recognition \cite{mishkin2015place}, etc., but very few research are conducted for using the content based image retrieval for localization purpose, especially indoor localization. One major reason is that to represent a specific location using images, all the visual information around this location have to be collected and stored to make this landmark distinguishable, which requires many images if using a normal Field Of View (FOV) camera, instead of a single image. Also, the features used for retrieving from the input image should depend on the location only, independent of the camera's orientation, but this is hard for normal FOV cameras, because the scene images are usually captured from different perspectives, and they therefore generate different features even though images are captured at the same location. 

Using GPU to accelerate image retrieving process is used by \cite{sattigeri2012implementation}, where each image is divided into patches and SIFT features are detected and indexed for each patch before stored into a database for retrieving. However, this work only uses a single GPU kernel and one-level parallel algorithm, which does not utilize the parallelism property of the data and is not efficient when the database scale increases. In our work,  a hierarchy multi-level parallel algorithm is used by identifying both the database and algorithm parrallelism property with multiple GPU kernels.

\section{ALGORITHMS}\label{Algorithm}

Before a localization system is available for usage, all the interested landmarks within the area where localization service is provided shall be modeled with the omnidirectional image features  and stored into the remote server database. Once this is done, the task of localizing an user's location using new captured images is  converted into a problem of finding the most similar image feature within the database. In this paper, we use the omnidirectional image feature proposed by \cite{hu2014mobile} to model the indoor environment, and the same type of feature extracted with the same procedure is later used for retrieval. Since for each landmark, an omnidirectional image can capture all the visual information around the user with single shot, each location can be modeled as one omnidirectional feature, instead of multiple features if using normal field of view camera \cite{6485272}. The FFT mangnitude of the one-dimensional omnidirectional vector is used as a rotation-invariant omnidirectional feature.

When the database scales up for a large scene, it is very time consuming to sequentially match the input frame features with all the images in the database of the large scene. So we use General Purpose GPU (GPGPU) to parallelize the query procedure and accelerate the querying speed. Section \ref{GPU_retrieving} will explain the algorithm of GPU-accelerated parallel feature retrieving. 

Since many man-made structures and objects look very similar, particularly for indoor environments, it is still a challenging task  for accurate localization using only one single image. We therefore adopt a multiple video frames querying mechanism by extracting a sequence of omnidirectional features from a short video clip, which greatly reduce the probability of false positive retrieving. Section \ref{aggregation} will discuss the multi-frame based candidates aggregation algorithm. 

\begin{figure}[t]
    \centering
    \fbox{\includegraphics[width=0.45\textwidth]{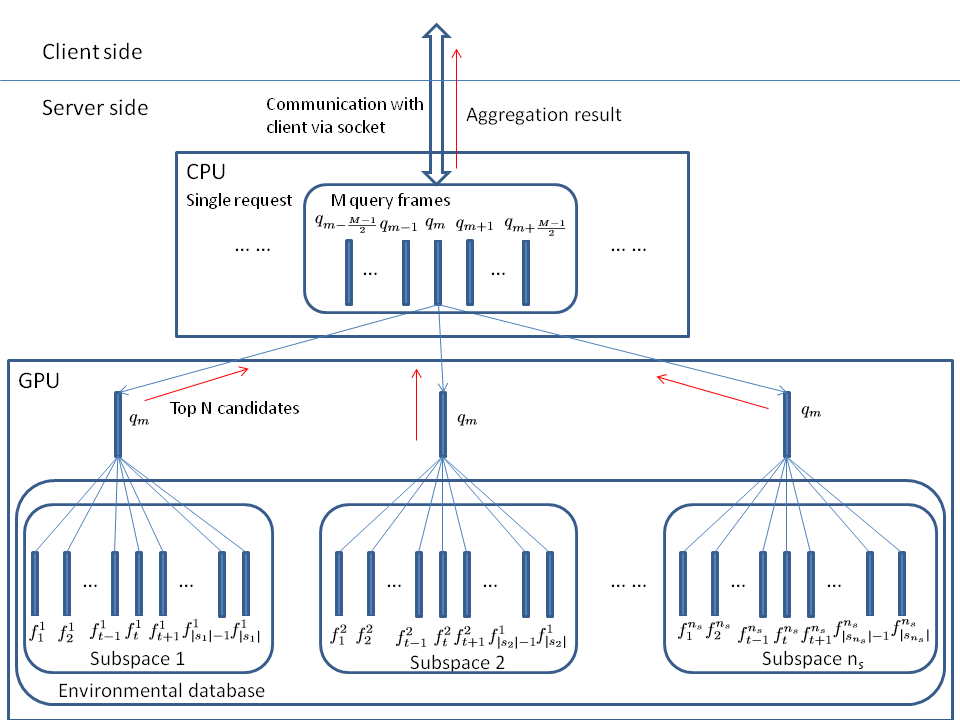}}
    \caption{Parallel retrieving with GPU in the remote server}
    \label{fig:Paralle_query}
\end{figure}

\subsection{GPU-accelerated parallel retrieval}\label{GPU_retrieving}
Without using GPU, the straightforward algorithm is to sequential compare the input feature vector with all the feature vectors in a database. This is problematic when the database scale increases, for example, to hundreds of thousands of features or more, and a real-time performance requirement is needed. We therefore design a parallel feature retrieving algorithm with NVIDIA GPU-acceleration. 

The retrieving process  have three levels of parallelisms. First, we divide the search database into multiple subspaces, for example, each floor data is counted as one subspace, and all the subspaces can be parallelly searched with CUDA streams. Second, within each subspace, we  process all the frames of an input query video clip in parallel instead of retrieving with each input frame one after another. Third,  we  use multiple threads to compare each input frame with multiple database frames simultaneously, rather than comparing with them one-by-one.

A diagram of this multi-level parallel retrieving process is illustrated in Fig. \ref{fig:Paralle_query}.  When a client sends a localization request of a frame $q_m$, it actually sends $\frac{M-1}{2}$ frames before the frame $q_m$, and $\frac{M-1}{2}$ frames after the frame $q_m$ to the server (e.g. $M$ = 3 or 5).  The CPU of the server will copy each frame's feature to the GPU, and receive the top $N$ candidates calculated by the GPUs for each frame after processing.  The database is divided into $n_s$ subspaces, and the $n_i$ subspace has $|n_i|$ modeled frames. All the subspaces are searched in parallel. Take the frame $q_m$ as an example. The frame $q_m$ is compared with all the frames $f_t^i$ ($t$ = 1, 2, ..., $|n_i|$) within each subspace $i$ ($i$ = 1, 2, ..., $n_s$). The same procedure is carried out to all frames. Finally, all the returned candidates of the $M$ query frames are returned to CPU for aggregation.

The detailed algorithm is shown in Algorithm  \ref{Algorithm_GPU}. In the initialization step, the server receives the data from the client and prepare for retrieving. In Step 2, for each retrieving, the server will not only select the  current testing frame, but also multiple other frames near the testing frame. Then, all the selected frames are copied to each CUDA stream one by one (Step 5), where the database of each subspace are stored and computation is taken care of concurrently. Within each stream, there are multiple threads, e.g. $32\times256$ threads, and each thread is responsible for the comparison of the querying frame and database frame (Step 7). Streams are executed simultaneously, and all the threads within each stream are also carried out in parallel. CUDA synchronization is called to make sure all the tasks within each thread (Step 9) and all the work in each stream (Step 12) are finished before the program moves to the next step. Finally, the GPU collects the returned candidates.

\begin{algorithm}[h]
    \SetInd{0.5em}{0.5em}
    \SetAlgoLined
    \KwData{M query frames received from CPU}
    \KwResult{Location candidates \textit{FinalResult} calculated by GPU}
    initialization\;
    selectNearbyFrames()\;
    \For{each selected frame $q_m$}
    {
        \For{each CUDA stream}
        {
            memoryCopyFrmoCPUtoGPU($q_m$)\;
            \For{each CUDA thread}
            {
                calculateDistance($q_m$, $f_i$);
            }
            synchronizeThreads()\;
        }
        synchronizeStream()\;
        \textit{Results} = selectTopCandidates()\;
        memoryCopyFromGPUtoCPU(\textit{Results})\;
    }       
    \caption{GPU multi-stream paralleling retrieving}
    \label{Algorithm_GPU}
\end{algorithm}

\subsection{Multi-frame candidate aggregation}\label{aggregation}
For many indoor places, such as narrow corridors, modeling the environment by video-taping a single walking is sufficient. However, to cover large areas, such as wide corridors or rooms, multiple paths of video capture are needed. Also, after the multi-frame retrieving process,  a large amount of position candidates are returned by the GPU, so a candidates aggregation algorithm is needed to aggregate all these candidates for a finalized location estimation to be delivered to the user. In an ideal case, all the candidates should be located at the position where the input image is captured. However, because of various reasons (noise, scene similarity, etc.), some of the candidates may be far away from the ground truth position, even though the majority of the candidates will cluster around the ground truth area. In this paper, we  design and implement a 2D multi-frame aggregation algorithm by utilizing the densities of the candidates' distribution, and calculate the most likely location estimation. The algorithm is shown in Algorithm \ref{Algorithm_aggregation}.

\begin{algorithm}[!t]
    \SetAlgoLined
    \SetInd{0.5em}{0.5em}
    \KwData{Location candidates on the floor plan coordinate system}
    \KwResult{Aggregated final location coordinate $FinalPosition$}
    initialization\;
    \For{each candidate $C_i$} 
    {
        Accumulate locationDistriArray($C_i$); 
    }       
    rankLocationDistribution(locationDistriArray)\;
    TopC = findTopDensityArea(locationDistriArray)\;
    \For{$index$ $<$ $NumOfTopC$}
    {
        \If{topDensityArea($index$) $>$ $NumOfAllCandidates \times tolerPer$ }
        {
            $FinalPosition$ = topDensityArea($index$)\;
            break;
        }
    }
    \caption{2D multi-frame localization candidates aggregation}
    \label{Algorithm_aggregation}
\end{algorithm}

In the initialization step, the candidates' indexes in the modeling images are mapped to their actual floor plan 2D coordinates by checking the geo-referenced mapping table pre-generated while modeling the environment. In our experiments, we discretize the floor plan coordinates by setting the $x$ (from west to east) and $y$ (from south to north) direction's units to be 30 cm, which is identical to ground tiles' width.  Then in Step 2, we count each tile's candidates number by a 2D binning operation, and obtain its location distribution array $locationDistriArray$. We sort this array in a descending order in Step 5, and obtain the first $TopC$ number (currently $TopC$ = 10) of tiles in Step 6. In a perfect case, the top one tile of these $TopC$ candidates should be the best estimated result since it has the largest amount of candidates dropped in. This is, however, not always the case, because the tile with the most candidates may be a false positive estimation, and its nearby tiles may have very few candidates. To increase the robustness of the estimation, in Step 7, we set a tolerant circle around each tile, with radius 3m in our experiment, and count the total number of candidates within this circle. If the amount is greater than a threshold percent--- $tolerPer$ (e.g. $tolerPer$ = 20\%) of the total amount of candidates, the corresponding tile is the final position, and is returned to the user.

\section{EXPERIMENTS}
 Experiments on a database of an eight-floor building are carried out to demonstrate the capacity of the proposed system, with real-time response (14fps) and robust localization results. In the experiment we will describe below, a whole floor within a campus building are modeled with multiple paths (5 parallel paths in this case), and test the parallel retrieving and aggregation algorithms for localization. We first capture video sequences of the entire floor along these five paths as the training databases, and then capture two other video sequences as the testing data sets.

For each testing frame, we query the entire training database and find the most similar stored features with the smallest Euclidean distance. Many scenes in the environment may be very similar,  so the returned feature with the smallest distance may not necessary be the real match. We then select the top N (N = 15 in this experiment) features as the candidates. Also, to solve the scene similarity problem, for each querying frame, we not only use itself as the query key, but we also use M-1 (M = 11) other frames, (M-1)/2 before and (M-1)/2 after this frame, as the query keys to generate candidates. We test all these frames within all the P (P = 5) paths, the frames of which are all labeled on the floor plan coordinate system, and retrieve all the candidates. Consequently, there are total $N \times  M  \times P$ candidates.

Fig.\ref{fig:long} shows an example of the result. The red stars are the 825 location candidates of a query frame and its related 10 frames. Note that several frames may return the same location, so each red star may represent more than one candidate. In the left image of Fig.\ref{fig:long}, the coordinate system has the same scale as the floor plan. Each unit in both horizontal and vertical directions corresponds 30cm in the physical space. The green star shows the ground truth location and the blue circle shows the final estimation of the testing frame. The distribution of the candidates is also illustrated in the right plot.

The $x$ and $y$ coordinates of right figure of Fig.\ref{fig:long} are the floor plan coordinates, and the height of each point represents the number of candidates falling into this position. Since all the modeling frames captured near the testing frame have similar contents and thus have alike features, the majority of the candidates drop near the location where the testing frame is captured. In this example, as shown in Fig.\ref{fig:long}, the testing frame's ground truth location is on the top left corner of the floor plan (the green star), as we can see, the majority of the candidates drop on the top corner area. The rightmost corner in the right figure of Fig.\ref{fig:long} also shows that the number of candidates (representing by the height) near the ground truth position is larger than the rest of places. A video demo can be found at this link \footnote {\url{https://youtu.be/gPGGcAfFsvY}} for the whole experiment to visualize the accuracy and robustness of the parallel retrieving and aggregation algorithm.

\section{Conclusion and Future Work}
This paper proposes a framework of converting the traditional visual information based indoor localization problem into a GPU-accelerated image and video retrieval problem. An omnidirectional image feature is used to model each environmental landmark, and stored in remote server for retrieving/localization. To solve scene similarity problem and ensure a real-time performance, we propose a multi-frame based retrieval technique and a 2D candidates aggregation algorithm. For the future work, multiple modality (e.g. Blue-tooth beacons + images) can be integrated together to resolve scene repetition problem. Also, more testing on a larger scale scene, for example, in a campus scale with multiple buildings, can be carried out for implementing a more practical localization solution.
%\end{document}  % This is where a 'short' article might terminate

%ACKNOWLEDGMENTS are optional
\section{Acknowledgments}
The authors would like to thank the US National Science Foundation (NSF) Emerging Frontiers in Research and Innovation (EFRI) Program for the support under Award No. EFRI 1137172. The authors would like to thank Jeury Mejia for his assistance in collecting data, developing mobile phone apps and running the experiments. Also thanks are given to Dr. Hao Tang at BMCC/CUNY for his valuable discussions.  

\bibliographystyle{abbrv}
\bibliography{sigproc}  % sigproc.bib is the name of the Bibliography in this case
% ACM needs 'a single self-contained file'!
%\balancecolumns

%\balancecolumns % GM June 2007
% That's all folks!
\end{document}